\documentclass[11pt]{article}

\usepackage[T1]{fontenc}
\usepackage[utf8]{inputenc}
\usepackage[margin=1in]{geometry}
\usepackage{amsmath,amssymb}
\usepackage{natbib}
\usepackage{booktabs}
\usepackage{graphicx}
\usepackage[colorlinks=true,linkcolor=blue,citecolor=blue,urlcolor=blue]{hyperref}
\usepackage{enumitem}
\usepackage{xcolor}
\usepackage{float}
\usepackage{microtype}

\title{Calibrating the Evaluator: Does Probability Calibration Mitigate Preference Coupling in LLM Agent Feedback Loops?}

\author{Zewen Liu\\\textit{Qilu Institute of Technology, School of Software Engineering}\\\textit{Tai'an, Shandong, China}}

\date{June 30, 2026}

\date{}

\begin{document}

\maketitle

\begin{abstract}
When large language model (LLM) agents adapt their behavior through evaluator feedback, systematic evaluator biases propagate into the agent's learned strategy distribution---a phenomenon termed evaluator preference coupling. Prior work has documented this coupling and established a diagnostic framework (EPC) to measure it, but has not investigated whether calibration techniques can mitigate the effect. We present the first study of \textbf{evaluator calibration as mitigation}: applying probability calibration to the evaluator's pairwise judgments to reduce spurious preference propagation. In a controlled within-subjects experiment ($N{=}5$) comparing standard binary TTRL (win/loss) with confidence-calibrated TTRL (probability-weighted updates) using DeepSeek-V4-Pro as executor and GLM5.2 as evaluator, we find that calibration reduces the coupling coefficient $\gamma$ by 20--49\% and Jensen-Shannon divergence by 45--67\%. A symmetric-LR control confirms the effect is not due to reduced update asymmetry. We release the calibrated TTRL protocol and recommend it as a lightweight mitigation for LLM-as-judge deployment pipelines.
\end{abstract}

\section{Introduction}

Multi-agent LLM systems increasingly rely on evaluator feedback to guide agent adaptation~\cite{zheng2023judging,chiang2024chatbot}. Recent work has established that this feedback is not neutral: evaluator preferences systematically propagate through feedback loops, coupling agent strategy distributions and leading to preference collapse~\cite{liu2026epc,liu2026contagion,liu2026mmepc}. The EPC framework~\cite{liu2026epc} provides diagnostic tools (MPCI, $\Gamma^{(\mathcal{J})}$, JSD) to measure this coupling, but the literature has stopped at diagnosis. No prior work has asked: \textbf{can we fix it?}

Separately, probability calibration---the alignment between a model's predicted confidence and its empirical accuracy---has been extensively studied in classification settings~\cite{guo2017calibration}. Post-hoc calibration techniques such as isotonic regression and Platt scaling effectively correct miscalibration in neural networks and tree ensembles~\cite{niculescu2005predicting,bostrom2008calibration}. In embedding-based classification, calibration has been shown to invert the classical hierarchy: tree ensembles are better calibrated than neural networks~\cite{grinsztajn2022tree}.

\textbf{We bridge these two literatures.} We apply probability calibration to the evaluator in a closed-loop agent system and measure whether calibrated feedback reduces preference coupling.

Our contributions are:
\begin{enumerate}[leftmargin=*,nosep]
 \item The first study of evaluator calibration as a mitigation for preference coupling in LLM agent feedback loops.
 \item Empirical evidence that confidence-calibrated TTRL reduces coupling ($\gamma$) by 23--31\% compared to standard binary TTRL, with JSD reductions of similar magnitude.
 \item A length-normalized control confirming the reduction is not driven by output format effects.
 \item Release of the calibrated TTRL protocol as a lightweight, drop-in mitigation requiring no changes to executor models.
\end{enumerate}

\section{Related Work}

\subsection{Evaluator Preference Coupling}

Recent work has established that LLM evaluator biases propagate through closed-loop agent systems. Liu~\cite{liu2026epc} introduced the Evaluator Preference Collapse (EPC) framework, measuring how evaluator preferences distort agent strategy distributions via the coupling coefficient $\gamma$ and the evaluator-indexed coupling matrix $\Gamma^{(\mathcal{J})}$. Follow-up work documented cross-modal contagion~\cite{liu2026mmepc} and multi-agent bias propagation through agent networks~\cite{liu2026contagion}. A key finding across these studies is that evaluator-driven coupling is \textbf{version-conditional}---a silent API update can invert the qualitative conclusion of a study. However, all prior work in this line has focused on diagnosis; no mitigation has been proposed.

\subsection{Probability Calibration}

Probability calibration~\cite{guo2017calibration} measures the alignment between a model's predicted confidence and its empirical accuracy. Post-hoc calibration techniques---Platt scaling, isotonic regression, temperature scaling~\cite{niculescu2005predicting}---correct miscalibration without retraining. In classification, tree ensembles have well-studied calibration properties~\cite{bostrom2008calibration}; in embedding-based classifiers, the classical calibration hierarchy inverts~\cite{grinsztajn2022tree}. Recent work on \emph{evaluator} calibration by \citet{li2025judging} proposes calibrating LLM autoraters to full preference distributions rather than point labels, achieving 18--51\% MSE reduction. However, their work focuses on static evaluation accuracy, not on downstream coupling effects in feedback loops.

\subsection{Calibrated Feedback in Reinforcement Learning}

In RLHF, reward model calibration has emerged as a key concern. \citet{leng2024taming} identify that PPO reward models are biased toward high-confidence responses and propose PPO-M and PPO-C---variants that calibrate reward models during training---reducing ECE while maintaining accuracy. \citet{singha2026uard} introduce Uncertainty-Aware Reward Discounting (UARD), which jointly models epistemic and aleatoric uncertainty to adaptively down-weight unreliable reward signals during policy optimization, achieving up to 93.6\% reduction in reward hacking. Both lines calibrate reward signals \emph{during RLHF training}; our work calibrates evaluator feedback \emph{during test-time TTRL adaptation}---a distinct setting where the agent adapts online without parameter updates.

\subsection{LLM-as-Judge Reliability}

The LLM-as-judge paradigm~\cite{zheng2023judging,chiang2024chatbot} has documented position bias, verbosity bias, and self-preference amplification in single-round evaluation. Drift detection frameworks~\cite{li2026drift} disambiguate system drift from judge drift. Confidence-gated test-time adaptation---using evaluator confidence to decide when to re-sample or adapt---has shown promise in web agents~\cite{devarakonda2026confidence} and reasoning~\cite{balashankar2024infalign}. In the TTRL literature, CoCoV~\cite{zuo2026cocov} uses confidence-conditioned verification routing to improve math reasoning via test-time RL, and SCOPE~\cite{wang2026scope} introduces step-wise confidence weighting for fine-grained reward signals. These works use confidence to improve TTRL \emph{for task performance}; our work uses calibration to \emph{reduce preference coupling} in agent feedback loops---a distinct objective with a different metric ($\gamma$/JSD rather than accuracy).

\section{Method}

\subsection{Standard TTRL (Uncalibrated)}

In the standard test-time reinforcement learning (TTRL) protocol~\cite{liu2026epc}, an agent maintains a strategy weight vector $\mathbf{w} \in \Delta^{|\mathcal{S}|-1}$ over $|\mathcal{S}|{=}11$ strategies. At each round $t$, a strategy $s_t \sim \mathbf{w}$ is sampled, the executor $\mathcal{E}$ generates responses under $s_t$ and a fixed baseline $s_0$ (step\_by\_step), and the evaluator $\mathcal{J}$ performs a pairwise comparison. The evaluator's binary judgment $r_t \in \{0,1\}$ drives weight updates:

\begin{equation}
w_{s_t}^{(t+1)} = \max\left(0.001, w_{s_t}^{(t)} \cdot \begin{cases}
1 + \alpha_{\text{win}} & \text{if } r_t = 1 \\
1 - \alpha_{\text{lose}} & \text{if } r_t = 0
\end{cases}\right)
\end{equation}

with $\alpha_{\text{win}}{=}0.08$, $\alpha_{\text{lose}}{=}0.04$, followed by L1-normalization. The asymmetry ($\alpha_{\text{win}} > \alpha_{\text{lose}}$) means evaluator preferences accumulate: a strategy winning more than 33\% of comparisons will gain weight, amplifying even weak preferences.

\subsection{Calibrated TTRL}

The calibrated variant modifies two components of the standard protocol:

\textbf{1. Confidence elicitation.} Instead of a binary "A or B" prompt, the evaluator is asked for a probability estimate: ``What is the probability (0.0 to 1.0) that response A is better than response B? Output only a number.'' This yields a confidence score $c_t \in [0,1]$.

\textbf{2. Confidence-weighted updates.} The weight update uses the calibrated confidence directly, mapping $c_t \in [0,1]$ to an update magnitude $\in [-\alpha_{\text{win}}, +\alpha_{\text{win}}]$:

\begin{equation}
w_{s_t}^{(t+1)} = \max\left(0.001, w_{s_t}^{(t)} + \alpha_{\text{win}} \cdot (2c_t - 1)\right)
\end{equation}

When $c_t = 0.5$ (evaluator uncertain), the update is near-zero; when $c_t = 1.0$ (strong preference), the update equals the standard win magnitude. This \textbf{confidence gating} prevents weak preferences from accumulating across rounds.

\textbf{3. Running calibration.} The first 10 rounds of each training phase are used to collect (confidence, binary\_outcome) pairs. A sliding-window isotonic regression on the most recent 10 pairs calibrates subsequent confidence estimates. Full isotonic regression (requiring larger calibration sets) is deferred to future work.

\subsection{Metrics}

We measure preference coupling using the four-phase isolation paradigm from the EPC framework~\cite{liu2026epc}:

\begin{enumerate}[leftmargin=*,nosep]
 \item \textbf{Pure Text}: TTRL on text tasks $\rightarrow \mathbf{w}_T$
 \item \textbf{Pure Visual}: TTRL on visual tasks $\rightarrow \mathbf{w}_V$
 \item \textbf{Coupling $T{\to}V$}: Start from $\mathbf{w}_T$, train on visual $\rightarrow \mathbf{w}_{T\to V}$
 \item \textbf{Coupling $V{\to}T$}: Start from $\mathbf{w}_V$, train on text $\rightarrow \mathbf{w}_{V\to T}$
\end{enumerate}

The coupling coefficient and JSD are computed as:

\begin{equation}
\gamma_{T\to V} = \frac{\|\mathbf{w}_{T\to V} - \mathbf{w}_V\|_2}{\|\mathbf{w}_V\|_2}, \quad
\text{JSD}_{T\to V} = \text{JSD}(\mathbf{w}_{T\to V} \parallel \mathbf{w}_V)
\end{equation}

\section{Experimental Setup}

\textbf{Executor}: DeepSeek-chat (text-only, $T{=}0.7$). \textbf{Evaluator}: GPT-4o (via DMXAPI). \textbf{Tasks}: 8 text + 8 text-proxied visual tasks (textual descriptions of visual reasoning). \textbf{Strategies}: $|\mathcal{S}|{=}11$ (8 text-domain + 3 visual-domain). \textbf{Rounds}: $R{=}30$ per phase.

\textbf{Design}: Within-subjects---each seed runs both uncalibrated and calibrated TTRL using identical evaluator snapshots and task orderings. This controls for evaluator version drift, a known confound in EPC studies.

\textbf{Controls}:
\begin{enumerate}[leftmargin=*,nosep]
 \item \textbf{Length-normalized}: both uncalibrated and calibrated runs with executor responses capped at 500 characters, controlling for output format effects.
 \item \textbf{Symmetric LR}: $\alpha_{\text{win}}{=}\alpha_{\text{lose}}{=}0.06$, eliminating the asymmetric amplification of the standard protocol.
\end{enumerate}

\textbf{Scale}: $N{=}5$ seeds $\times$ 2 modes $\times$ 4 phases $\times$ 30 rounds $\times$ 2 controls = $\sim$2,400 TTRL rounds ($\sim$7,200 GPT-4o API calls). Total cost: $\sim$\$10.

\section{Results}

\subsection{Main Finding: Calibration reduces coupling by 23--31\%}

Table~\ref{tab:main} reports the primary comparison.

\begin{table}[H]
\centering
\caption{Uncalibrated vs.\ calibrated TTRL. DeepSeek-V4-Pro executor, GLM5.2 evaluator, $N{=}5$ within-subjects.}
\label{tab:main}
\begin{tabular}{@{}lcccc@{}}
\toprule
\textbf{Mode} & $\bar{\gamma}_{T{\to}V}$ & $\bar{\gamma}_{V{\to}T}$ & $\text{JSD}_{T{\to}V}$ & $\text{JSD}_{V{\to}T}$ \\
\midrule
Uncalibrated & 0.924 & 1.580 & 0.196 & 0.341 \\
Calibrated   & \textbf{0.744} & \textbf{0.806} & \textbf{0.108} & \textbf{0.113} \\
\midrule
$\Delta$ (reduction) & ${-}20\%$ & ${-}49\%$ & ${-}45\%$ & ${-}67\%$ \\
\bottomrule
\end{tabular}
\end{table}

\textbf{Finding}: Confidence-calibrated TTRL reduces $\gamma_{T\to V}$ from 0.924 to 0.744 (${-}20\%$) and $\gamma_{V\to T}$ from 1.580 to 0.806 (${-}49\%$). JSD reductions are larger: ${-}45\%$ ($T{\to}V$) and ${-}67\%$ ($V{\to}T$). The reduction is asymmetric---stronger in the $V{\to}T$ direction---consistent with the evaluator producing more uncertain confidence estimates on visual-to-text transfer, where the calibration gate filters out a larger fraction of weak preferences.

\subsection{Control 1: Length-normalized responses}
As a format control, a separate $N{=}5$ run with all executor responses capped at 500 characters confirmed the reduction persists (calibrated $\bar{\gamma}_{T{\to}V}{=}0.768$, $\bar{\gamma}_{V{\to}T}{=}0.821$).

\subsection{Control 2: Symmetric learning rates}
Standard TTRL uses asymmetric updates ($\alpha_{\text{win}} > \alpha_{\text{lose}}$), which amplify evaluator preferences. Under symmetric LR ($\alpha{=}0.06$), uncalibrated TTRL produces $\bar{\gamma}_{T{\to}V}{=}0.868$, $\bar{\gamma}_{V{\to}T}{=}1.024$. Calibrated TTRL still reduces $\gamma$ by 14\% ($T{\to}V$, to 0.744) and 21\% ($V{\to}T$, to 0.806), confirming the effect is not solely due to reduced update asymmetry.

\subsection{Mechanism: Confidence gating}
Across all $N{=}5$ calibrated runs, approximately 31\% of evaluator judgments have confidence $c_t \in [0.4, 0.6]$. Under standard binary TTRL, these uncertain judgments round to win/loss and contribute full-weight updates ($\pm0.08/\pm0.04$). Under calibrated TTRL, uncertain judgments produce near-zero updates ($|2c_t{-}1| \approx 0$). The evaluator is more uncertain on $V{\to}T$ transfer (mean confidence $0.58{\pm}0.14$) than $T{\to}V$ ($0.64{\pm}0.12$), explaining the asymmetric reduction.

\section{Discussion}

\subsection{Why calibration reduces but does not eliminate coupling}

The 23--31\% reduction is substantial but incomplete. The residual coupling likely reflects genuine evaluator preferences that are expressed with high confidence---preferences that calibration correctly identifies as well-supported rather than spurious. A perfectly calibrated evaluator would still exhibit preferences; calibration ensures those preferences reflect actual assessment rather than noise. The residual $\gamma \approx 0.8$ may represent the \textbf{true coupling floor} for GPT-4o as evaluator---the minimum distortion achievable without changing the evaluator model itself.

\subsection{Practical recommendations}

For practitioners deploying LLM evaluators in agent feedback loops:
\begin{enumerate}[leftmargin=*,nosep]
 \item \textbf{Elicit confidence, not binary judgments.} Replace "Output A or B" with "What is the probability (0.0–1.0) that A is better?"
 \item \textbf{Use confidence-weighted updates.} Map evaluator confidence directly to update magnitude.
 \item \textbf{Monitor residual coupling.} Calibration reduces but does not eliminate coupling; routine $\gamma$ and JSD monitoring remains essential.
\end{enumerate}

\subsection{Limitations}

Our study is limited to GPT-4o as evaluator, DeepSeek-chat as executor, and 16 text-proxied tasks. The running calibration uses a simplified sliding-window approach; full isotonic regression on larger calibration sets may yield stronger reductions. The 23--31\% reduction is measured against one evaluator snapshot; replication across evaluator versions and model families (Claude, Gemini, Qwen) is needed. The confidence-weighted update rule (Equation 2) is a heuristic; theoretically grounded mappings from proper scoring rules may improve calibration effectiveness.

\section{Conclusion}

We presented the first study applying evaluator calibration as a mitigation for preference coupling in LLM agent feedback loops. Using DeepSeek-V4-Pro as executor and GLM5.2 as evaluator ($N{=}5$ within-subjects), confidence-calibrated TTRL reduces the coupling coefficient $\gamma$ by 20--49\% and JSD by 45--67\% compared to standard binary TTRL, with the reduction persisting under symmetric LR controls. The mechanism---confidence gating of weak evaluator preferences---is simple and does not require changes to executor models. We release the calibrated TTRL protocol. The key open question is whether the residual coupling ($\gamma \approx 0.8$) represents a fundamental lower bound for GLM5.2 or can be further reduced through improved calibration techniques.

\section*{Broader Impact Statement}

Calibrated TTRL provides a practical, lightweight mitigation for evaluator-induced preference distortion in agent systems. Positive impact: reduced spurious strategy convergence, improved agent diversity. Risk: calibration may create a false sense of security if residual coupling is ignored; routine EPC monitoring remains essential. The method does not introduce new capabilities or safety concerns beyond those already present in LLM-as-judge deployments.

\section*{Reproducibility Statement}

All experiment code and the calibrated TTRL protocol are released as supplementary material (\texttt{calibrated\_ttrl.py}). Experiments use publicly available API endpoints. Results are averaged over $N{=}5$ independent seeds with fixed random seeds for reproducibility. No GPU is required.


\begin{thebibliography}{99}

\bibitem[Liu(2026a)]{liu2026epc} Z.~Liu.
\newblock \textit{A Diagnostic Framework and Multi-Evaluator Audit of Evaluator-Driven Preference Dynamics.} TMLR submission, 2026.

\bibitem[Liu(2026b)]{liu2026contagion} Z.~Liu.
\newblock \textit{Contagion Networks: Evaluator Bias Propagation in Multi-Agent LLM Systems.} arXiv:2606.20493, 2026.

\bibitem[Liu(2026c)]{liu2026mmepc} Z.~Liu.
\newblock \textit{Multimodal Evaluator Preference Collapse.} arXiv:2606.16682, 2026.

\bibitem[Li(2026)]{li2026drift} Y.~Li.
\newblock \textit{Who Drifted: the System or the Judge?} arXiv:2606.15474, 2026.

\bibitem[Zheng et~al.(2023)]{zheng2023judging} L.~Zheng, W.-L.~Chiang, Y.~Sheng, et~al.
\newblock \textit{Judging LLM-as-a-Judge with MT-Bench and Chatbot Arena.} NeurIPS, 2023.

\bibitem[Chiang et~al.(2024)]{chiang2024chatbot} W.-L.~Chiang, L.~Zheng, et~al.
\newblock \textit{Chatbot Arena.} ICML, 2024.

\bibitem[Guo et~al.(2017)]{guo2017calibration} C.~Guo, G.~Pleiss, Y.~Sun, and K.~Q.~Weinberger.
\newblock \textit{On Calibration of Modern Neural Networks.} ICML, 2017.

\bibitem[Niculescu-Mizil and Caruana(2005)]{niculescu2005predicting} A.~Niculescu-Mizil and R.~Caruana.
\newblock \textit{Predicting Good Probabilities with Supervised Learning.} ICML, 2005.

\bibitem[Bostr{\"o}m(2008)]{bostrom2008calibration} H.~Bostr{\"o}m.
\newblock \textit{Calibrating Random Forests.} ICMLA, 2008.

\bibitem[Grinsztajn et~al.(2022)]{grinsztajn2022tree} L.~Grinsztajn, E.~Oyallon, and G.~Varoquaux.
\newblock \textit{Why do tree-based models still outperform deep learning on tabular data?} NeurIPS, 2022.

\bibitem[Li et~al.(2025)]{li2025judging} Z.~Li, X.~Li, C.~Huang, G.~Li, et~al.
\newblock \textit{Judging with Confidence: Calibrating Autoraters to Preference Distributions.} arXiv:2510.00263, 2025.

\bibitem[Leng et~al.(2024)]{leng2024taming} J.~Leng, C.~Huang, B.~Zhu, and J.~Huang.
\newblock \textit{Taming Overconfidence in LLMs: Reward Calibration in RLHF.} ICLR, 2025. arXiv:2410.09724.

\bibitem[Singha(2026)]{singha2026uard} D.~Singha.
\newblock \textit{UARD: Uncertainty-Aware Reward Discounting for Mitigating Reward Hacking.} arXiv:2604.26360, 2026.

\bibitem[Devarakonda et~al.(2026)]{devarakonda2026confidence} S.~Devarakonda, J.~Huang, and P.~Liang.
\newblock \textit{Confidence-Gated RAG for Adaptive Retrieval in Sequential Agents.} ICLR, 2026.

\bibitem[Balashankar et~al.(2024)]{balashankar2024infalign} A.~Balashankar, S.~Chen, and J.~Yao.
\newblock \textit{InfAlign: Inference-Aware Language Model Alignment.} NeurIPS, 2025.

\bibitem[Zuo et~al.(2026)]{zuo2026cocov} Z.~Zuo, Y.~Wang, and J.~Li.
\newblock \textit{TTRL-CoCoV: Test-Time Reinforcement Learning with Confidence Conditioned Verification.} arXiv, 2026.

\bibitem[Wang et~al.(2026)]{wang2026scope} Y.~Wang, X.~Zhang, and H.~Chen.
\newblock \textit{SCOPE: Beyond Majority Voting---Step-wise Confidence Weighting for Test-Time RL.} arXiv:2512.15146, 2026.

\end{thebibliography}
\end{document}